\documentclass[a4paper, conference]{IEEEtran}

\IEEEoverridecommandlockouts

\usepackage{mymacros}
\usepackage{times}
\usepackage{graphicx}
\usepackage{cite}

\usepackage[T1]{fontenc}
\usepackage{color}
\usepackage{hyperref}
\usepackage{pifont}

\usepackage{enumitem}

\begin{document}

\title{
Moment-based Adversarial Training for \\
Embodied Language Comprehension
}

\author{
\IEEEauthorblockN{Shintaro Ishikawa}
\IEEEauthorblockA{
Keio University \\
Kanagawa, Japan \\
Email: shin.0116@keio.jp}
\and
\IEEEauthorblockN{Komei Sugiura}
\IEEEauthorblockA{
Keio University \\
Kanagawa, Japan \\
Email: komei.sugiura@keio.jp}
}

\maketitle

\begin{abstract}
In this paper, we focus on a vision-and-language task in which a robot is instructed to execute household tasks. Given an instruction such as "Rinse off a mug and place it in the coffee maker," the robot is required to locate the mug, wash it, and put it in the coffee maker. This is challenging because the robot needs to break down the instruction sentences into subgoals and execute them in the correct order. On the ALFRED benchmark, the performance of state-of-the-art methods is still far lower than that of humans. This is partially because existing methods sometimes fail to infer subgoals that are not explicitly specified in the instruction sentences. We propose Moment-based Adversarial Training (MAT), which uses two types of moments for perturbation updates in adversarial training. We introduce MAT to the embedding spaces of the instruction, subgoals, and state representations to handle their varieties. We validated our method on the ALFRED benchmark, and the results demonstrated that our method outperformed the baseline method for all the metrics on the benchmark.
\end{abstract}

\section{Introduction}
\label{sec:introduction}

In our aging society, the need for daily care and support is increasing. Accordingly, the shortage of home care workers has become a social problem, and domestic service robots (DSRs) that can physically assist people with disabilities are expected to resolve it. Although natural language interfaces for DSRs are user-friendly, their ability to comprehend linguistic instructions regarding household tasks is still insufficient.

In this paper, we aim to build a multimodal language comprehension model that translates from language to sequences of actions.
Given an instruction such as ``Rinse off a mug and place it in the coffee maker,'' the robot should locate the mug, wash it, and put it in the coffee maker.

Users are apt to specify such household tasks with high-level instructions that are difficult for robots to decompose into subgoals. For the above example, the task consists of three subgoals, which should be executed in the correct order. Humans can easily accomplish these hierarchical tasks; however, it is challenging for robots to execute them correctly. For example, on the ALFRED benchmark, human performance was reported to be 91.0\% in unknown environments\cite{shridhar2020alfred}; however, state-of-the-art methods (e.g., FILM\cite{min2021film}) achieved less than 30\%.

Existing methods that use hierarchical structures attempt to generate subgoals to achieve high-level goals. However, it is sometimes difficult to predict subgoals when they are not specified explicitly in the high-level instructions.
Therefore, these models often fail to predict the correct subgoals, particularly in unknown environments, which leads to the robot failing to execute the overall tasks.

\begin{figure}[t]
    \centering
    \includegraphics[width=\linewidth]{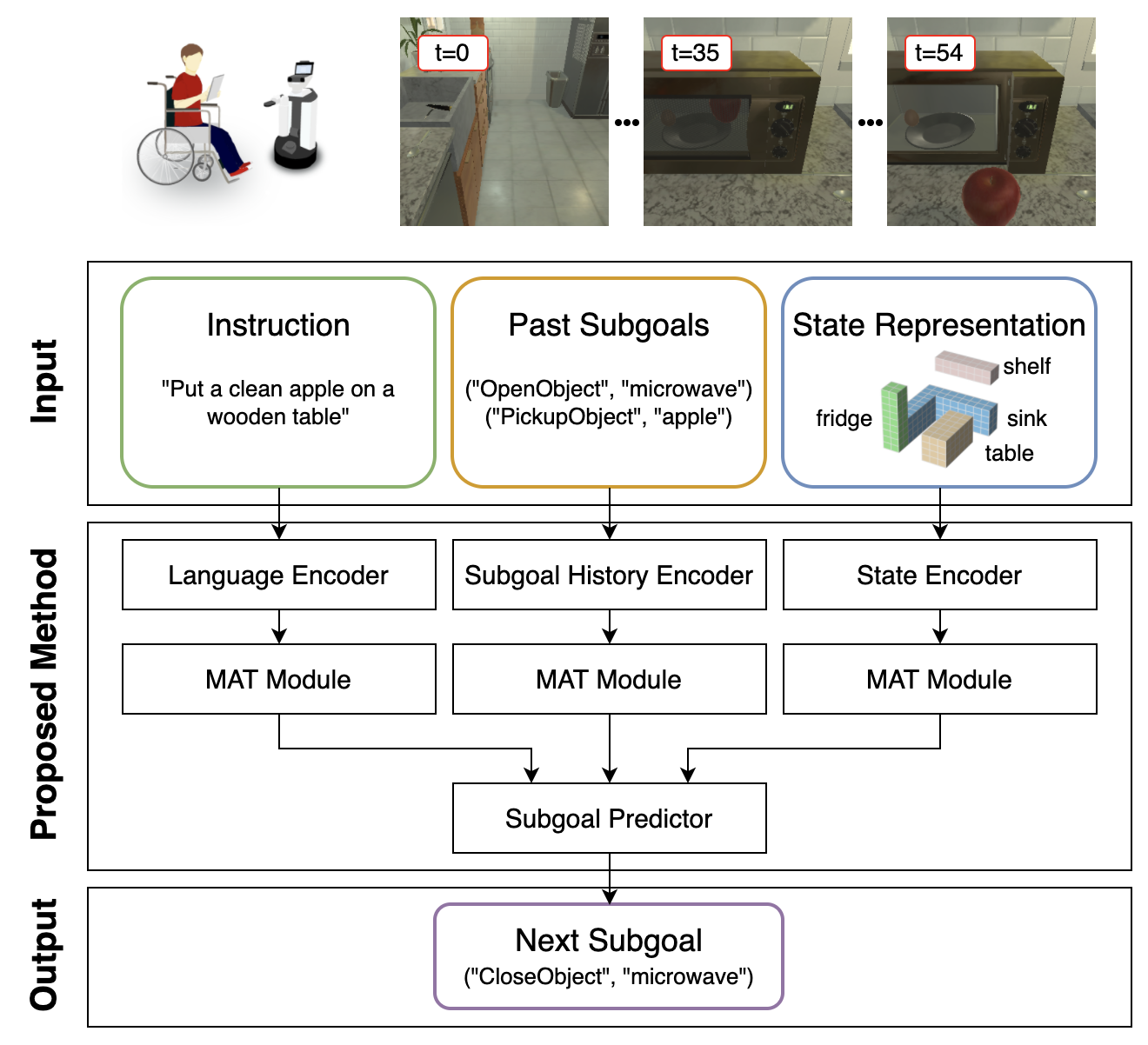}
    \vspace{-7mm}
    \caption{\small Our method overview: we add perturbations to each embedding space through the MAT modules.}
    \label{fig:overview}
    \vspace{-5mm}
\end{figure}

In this paper, we propose the introduction of adversarial training into the process of subgoal prediction to handle many different scenarios. We believe that this improves generalization performance and enables the robot to handle a variety of environments.
Fig.~\ref{fig:overview} shows an overview of our method.
The difference from existing methods is that our method is an adversarial approach based on VILLA\cite{gan2020large} with perturbations added to the embedding spaces of subgoals and state representations.
Our method can accomplish tasks accurately with general-purpose subgoal predictions as a result of adversarial training.
Our code is available at this URL\footnote{https://github.com/keio-smilab22/HLSM-MAT}.

The contributions of this study are summarized as follows:
\begin{itemize}
    \item We introduce adversarial perturbations to the embedding spaces of subgoals and state representations.
    \item We propose Moment-based Adversarial Training (MAT), which is a new algorithm that uses two types of moments for perturbation updates.
\end{itemize}

\section{Related Work}
\label{sec:related}

There have been many studies in the field of the robot-and-language studies such as Vision-and-Language Navigation (VLN)\cite{anderson2018vision,koller2010toward,tellex2011understanding,kuo2020deep,li2022reve}. \cite{wu2021visual} is a survey paper, which provides a comprehensive survey of VLN tasks and performs classification according to the characteristics of the language instructions in the tasks.
In the field, transformer-based models using attention mechanisms have recently achieved high performance.
MTCM\cite{magassouba2019understanding} is a model that specifies the target object from the instruction given a visual observation. \cite{ishikawa2021target} proposes a transformer-attention mechanism to model the relationships between objects and instructions for DSRs.

VLN tasks can be divided into two types: single-turn (e.g., \cite{MacMahon2006WalkTT}) and multi-turn (e.g., \cite{savva2019habitat}). For single-turn tasks, a series of instructions is provided before the robot starts to roll. According to whether a route is specified, instructions can be divided into goal-orientation (e.g., \cite{misra2018mapping}) and route-orientation (e.g., \cite{anderson2018vision}). The former contains several goals, but no clue about how to achieve them, and the latter contains specific trajectories. CrossMap Transformer\cite{magassouba2021crossmap} is a model that encodes linguistic and visual features to sequentially generate a path. It uses double back-translation to improve the mapping between linguistic and action features. FILM\cite{min2021film} is a modular method for embodied instruction following. To achieve the given goal, it builds a semantic metric map of the scene and performs exploration with a semantic search policy.

For multi-turn tasks, the instructions are given by a guide to a navigator in several turns. According to whether the navigator can respond to the guide, tasks are divided into imperative and interactive. In the former case, the navigator cannot respond to the guide and can only execute the instructions. In the latter case, both the guide and navigator can ask questions and share information. CVDN\cite{thomason2020vision} provides multi-turn interactive tasks with human-to-human dialogue situated in simulated, photo-realistic home environments.

Large-scale datasets are required to train the above models.
SUNCG\cite{song2017semantic} is a manually created dataset of 45,622 synthetic 3D scenes, ranging from a single room to multi-floor houses. The scenes include a variety of objects and furniture layouts. Matterport 3D\cite{chang2017matterport3d} contains 10,800 panoramic views from 194,400 RGB-D images of 90 building-scale scenes. The dataset is fully annotated with surface reconstructions, camera poses, and 2D and 3D semantic segmentation. Room-to-Room\cite{anderson2018vision} is a dataset for natural language navigation in real buildings. The environments are defined by navigation graphs, where nodes are viewpoints with a self-centered panoramic image, and edges signify robot-navigable transitions between two viewpoints.

Our method is inspired by the aforementioned VLN methods, such as HLSM\cite{blukis2021persistent}. HLSM is a representative hierarchical model and uses spatial representation as long-term memory to solve long-horizon tasks. It consists of a high-level controller that generates subgoals, and a low-level controller that generates sequences of actions to accomplish them. Our model is different from HLSM in that we take an adversarial approach by adding perturbations to the embedding spaces of subgoals and state representations.

\begin{figure}[t]
    \centering
    \includegraphics[width=\linewidth]{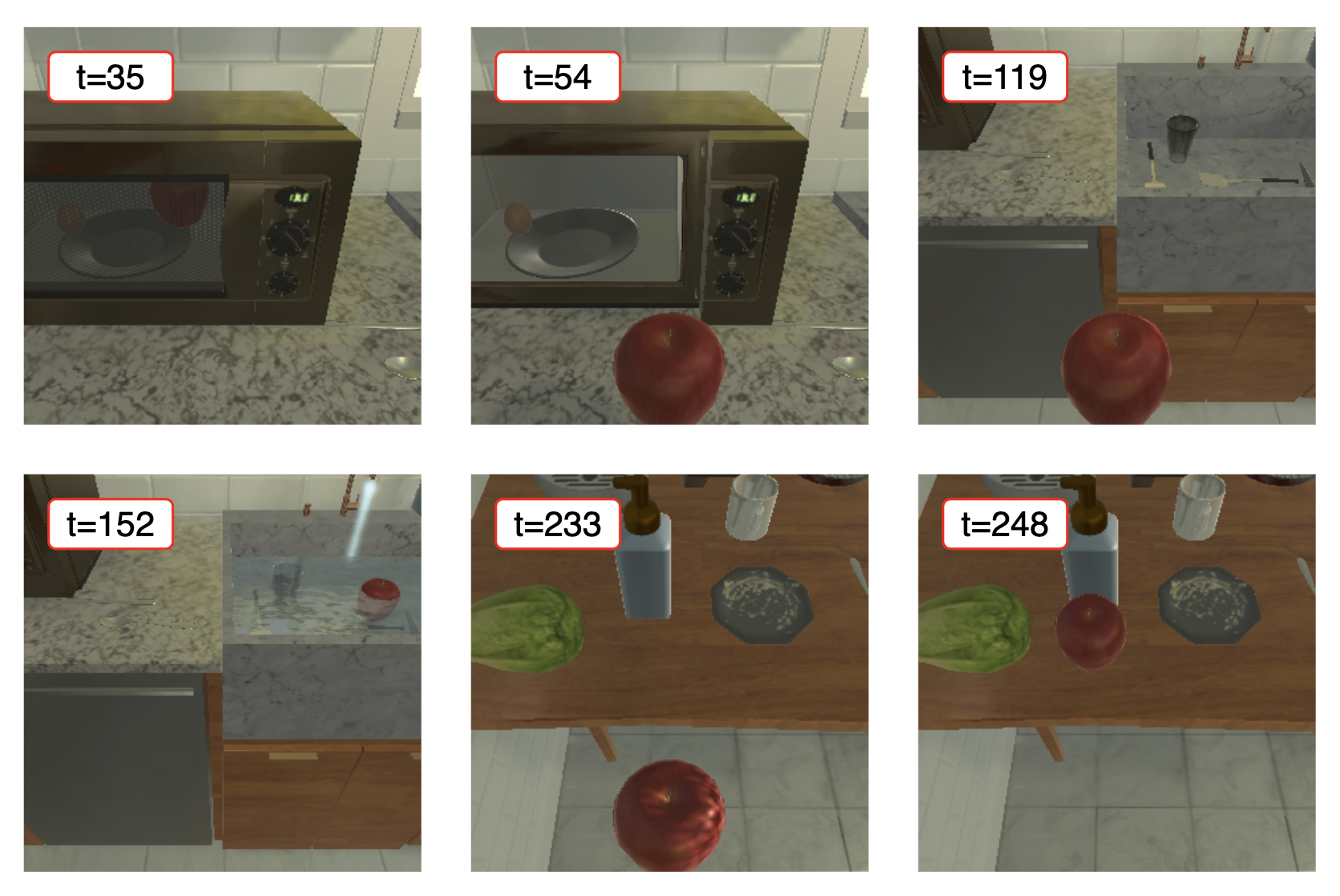}
    \vspace{-7mm}
    \caption{\small Typical scene in the ALFRED dataset. The instruction is ``Put a clean apple on a wooden table.'' The robot starts from the initial position (t=0). The robot should walk to the microwave (t=35) and pick up the apple from inside it (t=54), take the apple to the sink (t=119) to wash it (t=152), and then carry the clean apple to a wooden table (t=233) and put it on the table (t=248).}
    \label{fig:sample}
    \vspace{-5mm}
\end{figure}

\begin{figure*}[t]
    \centering
    \includegraphics[width=\linewidth]{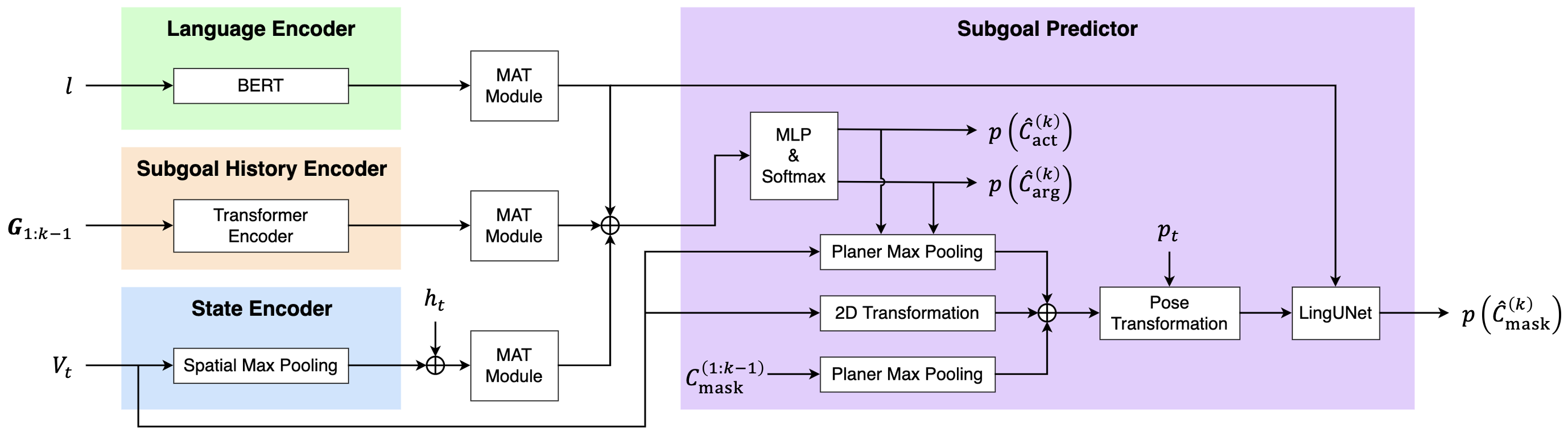}
    \vspace{-7mm}
    \caption{\small Our method framework: our method consists of a State Encoder, Language Encoder, Subgoal History Encoder, and Subgoal Predictor. Each module is explained in Section~\ref{sec:proposed}.}
    \label{fig:network}
    \vspace{-5mm}
\end{figure*}

\section{Problem Statement}
\label{sec:problem}

Our task is to build models that translate from language to sequences of actions and interactions in a simulated environment. We use the ALFRED benchmark\cite{shridhar2020alfred}, which includes long compositional goals in domestic environments.
The model should output appropriate actions at each timestep and achieve the given goal that consists of several household subgoals.
Fig.~\ref{fig:sample} shows a typical scene in the ALFRED dataset. For this scene, the goal is ``Put a clean apple on a wooden table.'' First, the robot is required to go to the microwave and pick up the apple from inside it. Second, the robot should take the apple to the sink and wash it. Finally, the robot should walk to the wooden table with the clean apple and put it on the table.
This task is characterized as follows:
\begin{itemize}
    \item \textbf{Input}: High-level instruction, and camera observations at each timestep.
    \item \textbf{Output}: Predicted probabilities that each action is executed at each timestep.
\end{itemize}

The terminology used in this paper is defined as follows:
\begin{itemize}
    \item \textbf{High-level instruction}: A natural language directive that the robot should accomplish throughout the task. The robot should divide it into subgoals that are not explicitly specified.
    \item \textbf{Subgoal}: A subtask that is composed of an interaction action and its target object.
    \item \textbf{Action}: One of the 13 atomic actions the robot can execute in the AI2-THOR environment\cite{kolve2017ai2}. The actions include ``MoveAhead,'' ``RotateRight,'' ``RotateLeft,'' ``LookUp,'' ``LookDown,'' ``PickUp,'' ``Put,'' ``Open,'' ``Close,'' ``ToggleOn,'' ``ToggleOff,'' ``Slice,'' and ``Stop.''
\end{itemize}

In this paper, we assume that the robot handles household tasks using a sequence of discrete actions in a simulated environment. We use the AI2-THOR simulator, which is a near photo-realistic interactable framework for robots in a domestic environment.
We use a simulation-based dataset to make the experimental results reproducible and improve the robot’s performance at a low cost.
In the ALFRED benchmark, there are eight evaluation metrics explained in \cite{shridhar2020alfred}. Among them, we use Seen/Unseen Success Rate (SR) and Seen/Unseen Goal-Conditioned Success Rate (GC) to evaluate our model. We describe these metrics in detail in Section~\ref{sec:experiments}.

\section{Proposed Method}
\label{sec:proposed}

Fig.~\ref{fig:network} shows the framework of our method. Our method consists of four modules: State Encoder, Language Encoder, Subgoal History Encoder, and Subgoal Predictor, and each module is explained below. The household tasks on VLN benchmarks such as ALFRED\cite{shridhar2020alfred} sometimes have hierarchical goals that are specified using high-level directives. Therefore, building a hierarchical model to handle the tasks is a reasonable approach. Our method is inspired by VLN methods such as HLSM\cite{blukis2021persistent}, which is a representative hierarchical model. Although the proposed model is based on HLSM, our approach is also applicable to other VLN methods, such as Episodic Transformer\cite{pashevich2021episodic}.

\subsection{Input}
At timestep $t$ with the $k$-th subgoal prediction, the input $\bm{x}_{t,k}$ to our model is defined as follows:
\begin{align}
    \bm{x}_{t,k}&=\{l,\bm{G}_{1:k-1},\bm{s}_t\}, \\
    \bm{G}_{1:k-1}&=\{\bm{g}_1,\bm{g}_2,\cdots,\bm{g}_i,\cdots,\bm{g}_{k-1}\}, \\
    \bm{g}_i&=\{C^{(i)}_{\mathrm{act}},C^{(i)}_{\mathrm{arg}},C^{(i)}_{\mathrm{mask}}\}, \\
    \bm{s}_t&=\{V_t,h_t,p_t\},
\end{align}
where $l$, $\bm{G}_{1:k-1}$, and $\bm{s}_t$ denote the instruction, sequence of past subgoals, and current state representation, respectively. $C^{(i)}_{\mathrm{act}}$, $C^{(i)}_{\mathrm{arg}}$, and $C^{(i)}_{\mathrm{mask}}$ denote the interaction type (e.g., ``PickUp'' and ``Open''), semantic class of the interaction argument (e.g., ``Apple'' and ``Microwave''), and 3D mask that identifies the location of the argument instance, respectively. $\bm{s}_t$ captures the cumulative robot knowledge of the world, and is generated in the observation model proposed in \cite{blukis2021persistent}. $V_t$, $h_t$, and $p_t$ denote the 3D voxel map that indicates the object classes in each voxel, 1-of-K vector that indicates the class of the object held by the robot, and pose of the robot, respectively. Note that the pose includes the 2D position, pitch angle, and yaw angle of the robot.

\subsection{Encoding Instructions, Subgoals and State representations}
The Language Encoder encodes the instruction $l$ using a pretrained BERT\cite{devlin2018bert} that we fine-tune during training. We use the CLS token embedding as the task embedding $\bm{\phi}_l$.
The Subgoal History Encoder encodes the sequence of past subgoals $\bm{G}_{1:k-1}$. We use a two-layer transformer autoregressive encoder to compute $\bm{\phi}^{(1:k-1)}_g$ and consider $\bm{\phi}^{(k-1)}_g$ as the subgoal history embedding vector.
The State Encoder encodes the current state representation $\bm{s}_t$ as $\bm{\phi}^{(t)}_s=\{h_t,f_{\mathrm{mp\_s}}(V_t)\}$ where $f_{\mathrm{mp\_s}}$ denotes a max-pooling operation over spatial dimensions, which transforms $V_t$ into a one-hot vector that represents the object classes in the observed space.

In the Language Encoder, we use the transformer attention mechanism to model the relationships between words. We use the same structure in the State Encoder to model the relationship between subgoals at different timesteps.

\subsection{Subgoal Prediction}
The Subgoal Predictor takes the concatenated representations $\bm{\Phi}_{t,k}=\{\bm{\phi}_l,\bm{\phi}^{(k-1)}_g,\bm{\phi}^{(t)}_s\}$ as the input. We obtain the subgoal type and argument class as follows:
\begin{align}
    p(\hat{C}^{(k)}_{\mathrm{act}})&=\mathrm{softmax}(f_{\mathrm{mlp}}(\bm{\Phi}_{t,k})), \\
    p(\hat{C}^{(k)}_{\mathrm{arg}})&=\mathrm{softmax}(f_{\mathrm{mlp}}(\bm{\Phi}_{t,k})),
\end{align}
where $f_{\mathrm{mlp}}$ denotes a multi-layer perceptron.

Next, we compute a bird's-eye view representation $\bm{b}_t$ as follows:
\begin{align}
    \bm{b}_{t,k}=\{a_t,\sum^{k-1}_{i=0}f_{\mathrm{mp\_p}}(C^{(i)}_{\mathrm{mask}}),f_{\mathrm{mp\_p}}(V_t)\},
\end{align}
where $a_t$ denotes a top-down view that represents each position with one or more of seven affordance classes: ``pickable,'' ``receptacle,'' ``togglable,'' ``openable,'' ``ground,'' ``obstacle,'' and ``observed.'' Additionally, $f_{\mathrm{mp\_p}}$ denotes a max-pooling operation over planar dimensions.

Finally, we compute the 3D argument mask as follows:
\begin{align}
    p(\hat{C}^{(k)}_{\mathrm{mask}})=f_{\mathrm{lun}}(f_{\mathrm{trans}}(\bm{b}_{t,k},p_t),\bm{\phi}_l),
\end{align}
where $f_{\mathrm{lun}}$ denotes a neural network based on the LingUNet architecture\cite{misra2018mapping} and $f_{\mathrm{trans}}$ performs the transformation from the bird's-eye view to the robot’s egocentric pose.
The final output $\bm{g}_k$ is given by $\bm{g}_k=\{C^{(k)}_{\mathrm{act}},C^{(k)}_{\mathrm{arg}},C^{(k)}_{\mathrm{mask}}\}$.

\subsection{Moment-based Adversarial Training}
Inspired by VILLA\cite{gan2020large}, we add adversarial perturbations in the feature space directly. 
The perturbation $\bm{\delta}$ is defined as $\bm{\delta}=\{\delta_l,\delta_g,\delta_s\}$ where $\delta_l$, $\delta_g$, and $\delta_s$ denote the perturbations for the instruction, sequence of past subgoals, and current state representation, respectively.

First, we compute the gradient of the cross-entropy loss w.r.t. $\bm{\delta}$.
\begin{align}
    E(\bm{\delta})&=\mathrm{CE}(f(\bm{x}+\bm{\delta}),\bm{g}), \\
    \nabla_{\bm{\delta}}E(\bm{\delta})&=\frac{\partial E}{\partial\bm{\delta}},
\end{align}
where $f$ and $\mathrm{CE}(\cdot,\cdot)$ denote the overall network and the cross-entropy loss, respectively.

Next, we introduce two types of Adam-like moving averages as follows:
\begin{align}
    \bm{m}_t&=\rho_1\bm{m}_{t-1}+(1-\rho_1)\nabla_{\bm{\delta}}E(\bm{\delta}_t), \\
    \bm{v}_t&=\rho_2\bm{v}_{t-1}+(1-\rho_2)(\nabla_{\bm{\delta}}E(\bm{\delta}_t))^2,
\end{align}
where $t$ denotes the current adversarial training step, and $\rho_1$ and $\rho_2$ are smoothing coefficients.
We obtain the perturbation update $\Delta\bm{\delta}_t$ as follows:
\begin{align}
    \hat{\bm{m}}_t&=\frac{\bm{m}_t}{1-(\rho_1)^t}, \\
    \hat{\bm{v}}_t&=\frac{\bm{v}_t}{1-(\rho_2)^t}, \\
    \Delta\bm{\delta}_t&=\eta\frac{\hat{\bm{m}}_t}{\sqrt{\hat{\bm{v}}_t+\varepsilon}},
\end{align}
where $\eta$ and $\varepsilon$ denote the learning rate in MAT and a small positive value, respectively.

Finally, we update the perturbation using $\Delta\bm{\delta}_t$.
\begin{align}
    \bm{\delta}_{t+1}=\Pi_{\|\bm{\delta}\|\leq\epsilon}(\bm{\delta}_t+\frac{\Delta\bm{\delta}_t}{\|\Delta\bm{\delta}_t\|_F}),
\end{align}
where $\Pi_{\|\cdot\|\leq\epsilon}$ performs a projection onto the $\epsilon$-ball and $\|\cdot\|_F$ denotes the Frobenius norm.

\subsection{Loss Function}
We use the following loss function.
\begin{align}
    L&=L_{cln}+L_{at}+\lambda L_{kl}, \\
    L_{cln}&=\mathrm{CE}(f(\bm{x}_{t,k}),\bm{g}_k), \\
    L_{at}&=\mathrm{CE}(f(\bm{x}_{t,k}+\bm{\delta}),\bm{g}_k), \\
    L_{kl}&=\mathrm{D}_{KL}(f(\bm{x}_{t,k})\|f(\bm{x}_{t,k}+\bm{\delta}))\nonumber \\
    &\qquad+\mathrm{D}_{KL}(f(\bm{x}_{t,k}+\bm{\delta})\|f(\bm{x}_{t,k})),
\end{align}
where $\lambda$ and $\mathrm{D}_{KL}(\cdot\|\cdot)$ denote the loss weight and Kullback--Leibler divergence, respectively.

\section{Experiments}
\label{sec:experiments}

\subsection{Experimental Setup}
The experimental setup is summarized in Table~\ref{tab:params}. Note that \#L, \#H, and \#A denote the number of layers, hidden size, and number of attention heads in each transformer, respectively. Adversarial training step $\alpha$ denotes the number of perturbation updates in each iteration. We set the batch size to 1 because one episode consists of many action steps.

Our model had 111 M parameters.
We trained the parameters on four Tesla V100s with 32GB of GPU memory.
It took 2 days to train our model and 2 minutes to infer the actions of one episode.

\begin{table}[H]
    \small
    \vspace{-2mm}
    \caption{Parameter settings and structures}
    \vspace{-2mm}
    \label{tab:params}
    \centering
    \begin{tabular}{l l l l}
        \hline
        Language Encoder & \#L: 12 & \#H: 768 & \#A: 12 \\
        \hline
        Subgoal History Encoder & \#L: 2 & \#H: 128 & \#A: 8 \\
        \hline
        Optimizer & \multicolumn{3}{l}{Adam ($\beta_{1}=0.9,\beta_{2}=0.999$)} \\
        \hline
        Learning rate & \multicolumn{3}{l}{$5\times10^{-5}$} \\
        \hline
        Epoch & \multicolumn{3}{l}{6} \\
        \hline
        Batch size & \multicolumn{3}{l}{1} \\
        \hline
        Loss weight & \multicolumn{3}{l}{1.5} \\
        \hline
        & \multicolumn{3}{l}{$\alpha=7$} \\
        MAT & \multicolumn{3}{l}{$\eta=1\times10^{-3}$} \\
        & \multicolumn{3}{l}{$\rho_{1}=0.9,\rho_{2}=0.999$} \\
        \hline
    \end{tabular}
    \vspace{-2mm}
\end{table}

\begin{table}[b]
    \small
    \vspace{-5mm}
    \caption{Dataset split}
    \vspace{-2mm}
    \label{tab:split}
    \centering
    \begin{tabular}{l c c c c c}
        \hline
        & \bf{Train} & \multicolumn{2}{c}{\bf{Validation}} & \multicolumn{2}{c}{\bf{Test}} \\
        & & Seen & Unseen & Seen & Unseen \\
        \hline
        \#Annotations & 21023 & 820 & 821 & 1533 & 1529 \\
        \#Scenes & 108 & 88 & 4 & 107 & 8 \\
        \hline
    \end{tabular}
\end{table}
\begin{table*}[t]
    \small
    \caption{Quantitative results and ablation studies}
    \vspace{-2mm}
    \label{tab:quantitative_results}
    \centering
    \begin{tabular}{l c c c c c c c c c c c c c}
        \hline
        & & & & & & \multicolumn{4}{c}{\bf{Validation}} & \multicolumn{4}{c}{\bf{Test}} \\
        \bf{Method} & \bf{Condition} & \bf{MAT} & \multicolumn{3}{c}{\bf{Perturbation}} & \multicolumn{2}{c}{Unseen} & \multicolumn{2}{c}{Seen} & \multicolumn{2}{c}{Unseen} & \multicolumn{2}{c}{Seen} \\
        & & & $\bm{\phi}_l$ & $\bm{\phi}_g$ & $\bm{\phi}_s$ & SR & GC & SR & GC & SR & GC & SR & GC \\
        \hline
        HiTUT G-only\cite{zhang2021hierarchical} & & & & & & 10.23 & 20.71 & 18.41 & 25.27 & 11.12 & 17.89 & 13.63 & 21.11 \\
        LAV\cite{nottingham2021modular} & & & & & & - & - & 12.7 & 23.4 & 6.3 & 17.3 & 13.4 & 23.2 \\
        HLSM\cite{blukis2021persistent} & & & & & & 18.28 & 31.24 & 29.63 & 38.74 & 20.27 & 30.31 & 29.94 & 41.21 \\
        \hline
        & full$^\dag$ & \checkmark & \checkmark & \checkmark & \checkmark & \bf{18.39} & 31.32 & 30.00 & 41.39 & 21.39 & 32.14 & 31.83 & 43.88 \\
        & full & \checkmark & \checkmark & \checkmark & \checkmark & 17.66 & 31.79 & 30.98 & 42.29 & \bf{21.84} & 32.41 & \bf{33.01} & 43.65 \\
        Ours & (i) & & \checkmark & \checkmark & \checkmark & 16.69 & 31.18 & \bf{31.71} & 41.82 & 20.99 & 32.03 & 30.66 & 43.10 \\
        & (ii)-a & \checkmark & & \checkmark & \checkmark & 15.71 & 30.66 & 30.85 & 41.54 & 21.39 & 31.03 & 32.29 & 43.30 \\
        & (ii)-b & \checkmark & \checkmark & & \checkmark & 15.96 & \bf{31.89} & 30.24 & \bf{42.63} & 21.52 & \bf{33.43} & 31.57 & \bf{44.38} \\
        & (ii)-c & \checkmark & \checkmark & \checkmark & & 14.25 & 29.06 & 30.73 & 40.68 & 20.27 & 30.66 & 30.14 & 41.23 \\
        \hline
    \end{tabular}
    \vspace{-5mm}
\end{table*}

\subsection{Dataset}
In the experiment, we evaluated our method on the ALFRED benchmark\cite{shridhar2020alfred}.
The ALFRED dataset consists of instructions corresponding to expert demonstration episodes. For every expert demonstration, instructions were collected from at least three annotators using Amazon Mechanical Turk.
We used this dataset for the evaluations because ALFRED is the standard VLN benchmark for household tasks in near photo-realistic scenes. It includes high-level and low-level instructions for object and environment interactions.
As preprocessing, we extracted subgoal sequences from each interaction action in the expert demonstrations. For every piece of the subgoal sequence, we computed the argument class $C_{\mathrm{arg}}$ and 3D mask $C_{\mathrm{mask}}$ using the mask in the egocentric visual observations.

The ALFRED dataset includes 25,743 English instructions describing 8,055 expert demonstrations, averaging 50 steps each. The expert demonstrations are classified into seven types of tasks: ``Pick \& Place,'' ``Stack \& Place,'' ``Pick Two \& Place,'' ``Clean \& Place,'' ``Heat \& Place,'' ``Cool \& Place,'' and ``Examine in Light.''

Table~\ref{tab:split} shows the dataset split. We followed the split defined in \cite{shridhar2020alfred}. The validation and test sets were split into seen and unseen folds. Scenes in the seen folds were subsets of those in the training fold; scenes in the unseen folds were separated from the training fold and from each other.
We used the training set to update our model’s parameters and tune the hyperparameters. We evaluated our model on both the validation and test sets.

\begin{figure}[b]
    \centering
    \vspace{-5mm}
    \includegraphics[width=\linewidth]{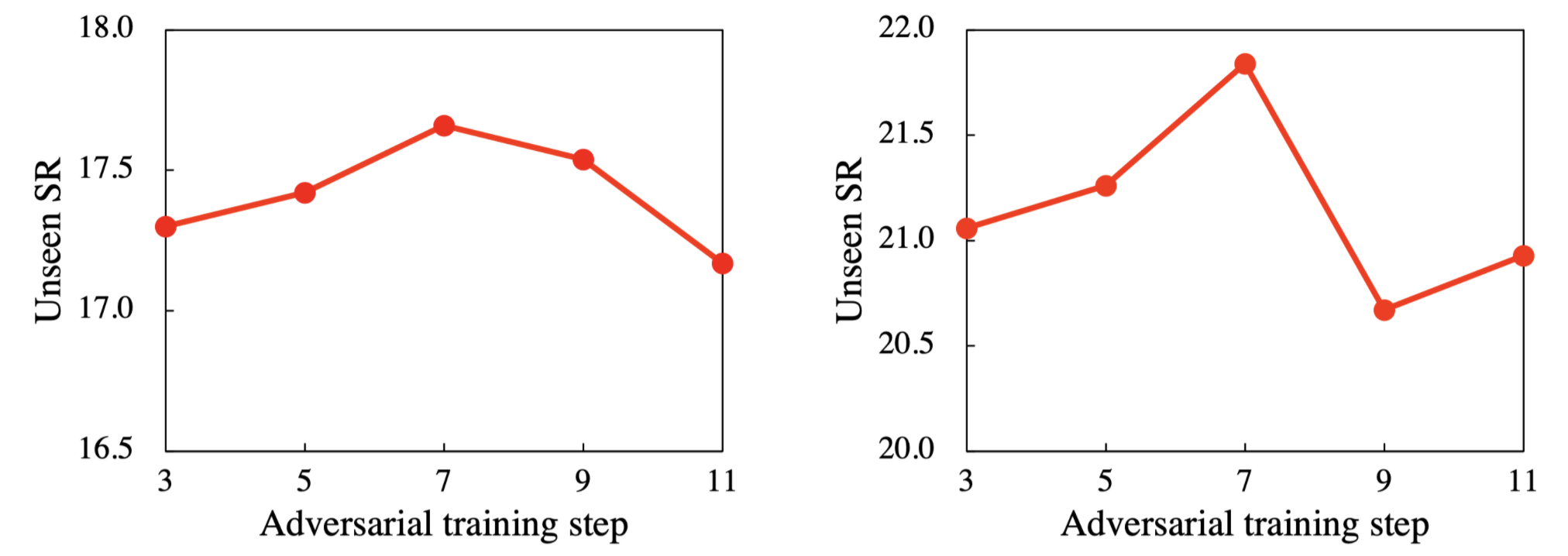}
    \vspace{-7mm}
    \caption{\small Effect of the adversarial training step. Our method achieved the highest score for an adversarial training step of 7 on both the validation and test sets.}
    \label{fig:graph}
\end{figure}

\subsection{Quantitative Results}
Table~\ref{tab:quantitative_results} shows the quantitative results.
We selected Seen/Unseen SR and Seen/Unseen GC as metrics following the standard evaluation method of the task.

SR is defined as $\mathrm{SR}=\frac{N_\mathrm{s}}{N}$ where $N$ and $N_\mathrm{s}$ denote the total number of episodes and number of episodes where the robot succeeds in the task, respectively. We consider a result as ``success'' if the object positions and state changes correspond correctly to the task goal conditions at the end of the action sequence.
We use the Unseen SR as a primary metric to evaluate our model.

GC is defined as $\mathrm{GC}=\frac{1}{N}\sum_{i=1}^N\frac{G^{(i)}_\mathrm{s}}{G^{(i)}}$ where $G^{(i)}$ and $G^{(i)}_\mathrm{s}$ denote the total number of subgoals in the $i$-th episode and number of completed subgoals in the $i$-th episode, respectively.

We selected HLSM as the baseline method because it is a representative hierarchical model and has been successfully applied to the ALFRED benchmark.
For the primary metric Unseen SR, our method achieved 18.39\% on the validation set, whereas the baseline method achieved 18.28\%. Additionally, our method achieved 21.39\% for the Unseen SR on the test set, whereas the baseline method achieved 20.27\%.
Our method outperformed HLSM by 1.11\% and 1.12\% on the validation and test sets, respectively. Note that, on the condition ``full$^\dag$,'' we applied MAT to a low-level controller as well as a high-level controller.

\begin{figure}[b]
    \centering
    \vspace{-5mm}
    \includegraphics[width=\linewidth]{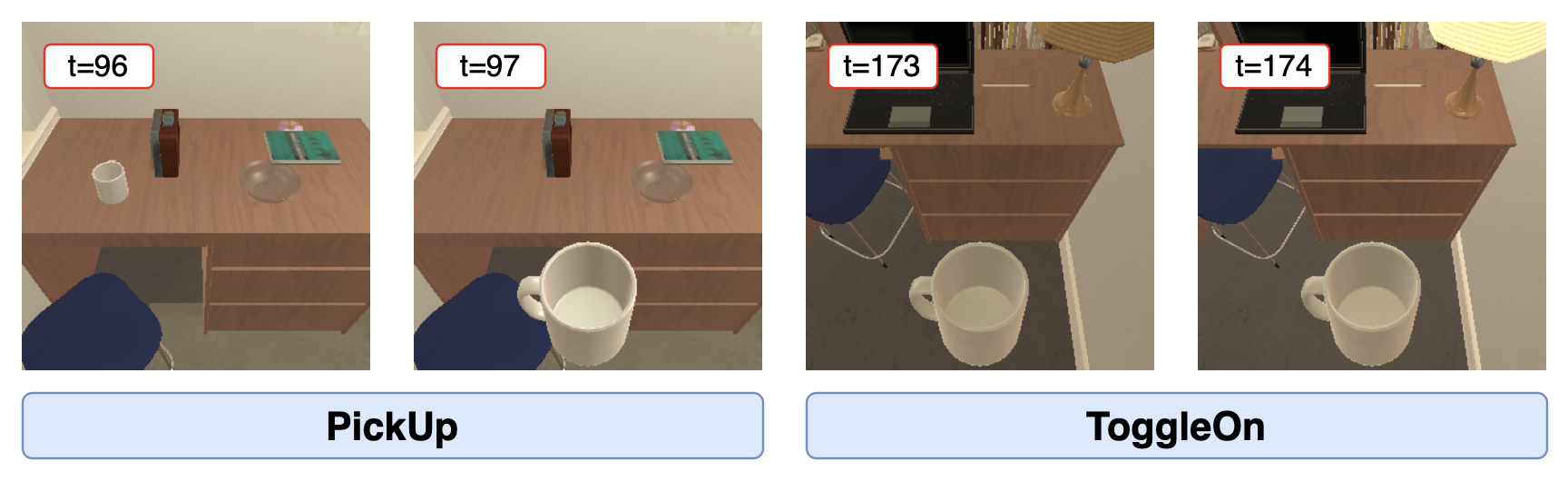}
    \vspace{-7mm}
    \caption{\small Success episode. The instruction is ``Examine a cup under a lamp.'' Snapshots of every interaction action taken during the placing task are shown.}
    \label{fig:success_01}
\end{figure}

\begin{figure*}[t]
    \centering
    \includegraphics[width=\linewidth]{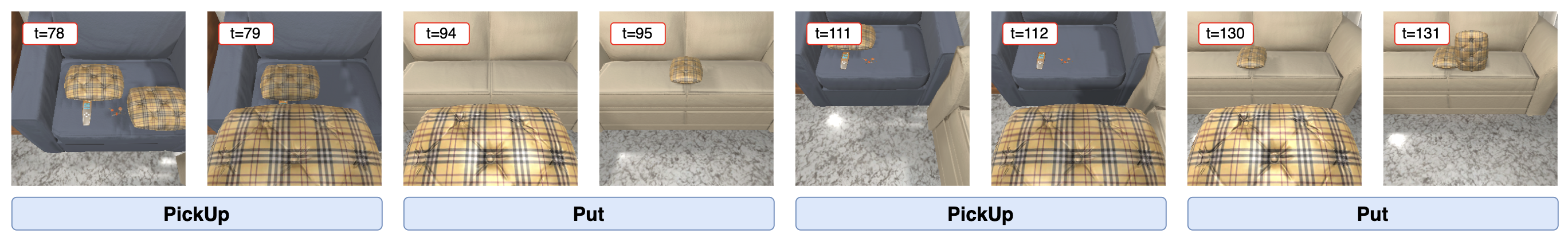}
    \vspace{-7mm}
    \caption{\small Success episode. The instruction is ``Place the two pillows on the sofa.''}
    \label{fig:success_02}
    \vspace{-2mm}
\end{figure*}

\begin{figure*}[t]
    \centering
    \includegraphics[width=\linewidth]{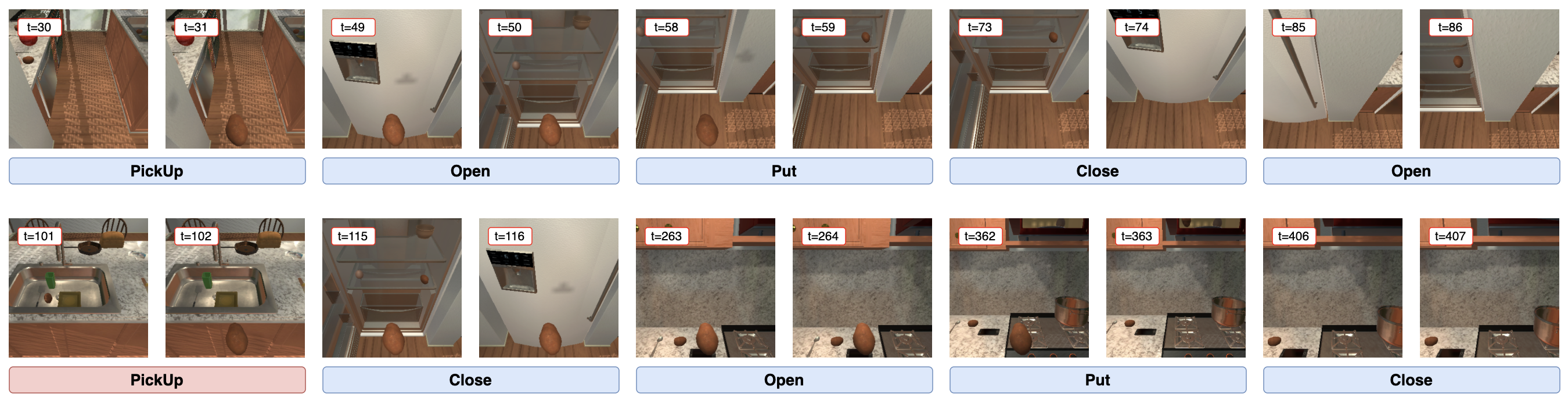}
    \vspace{-7mm}
    \caption{\small Failure episode. The instruction is ``Put a cooled potato inside the microwave.'' The robot succeeded in most of the subgoals; however, it was not able to pick up the cooled potato.}
    \label{fig:failure_01}
    \vspace{-5mm}
\end{figure*}

\begin{figure}[b]
    \centering
    \vspace{-5mm}
    \includegraphics[width=\linewidth]{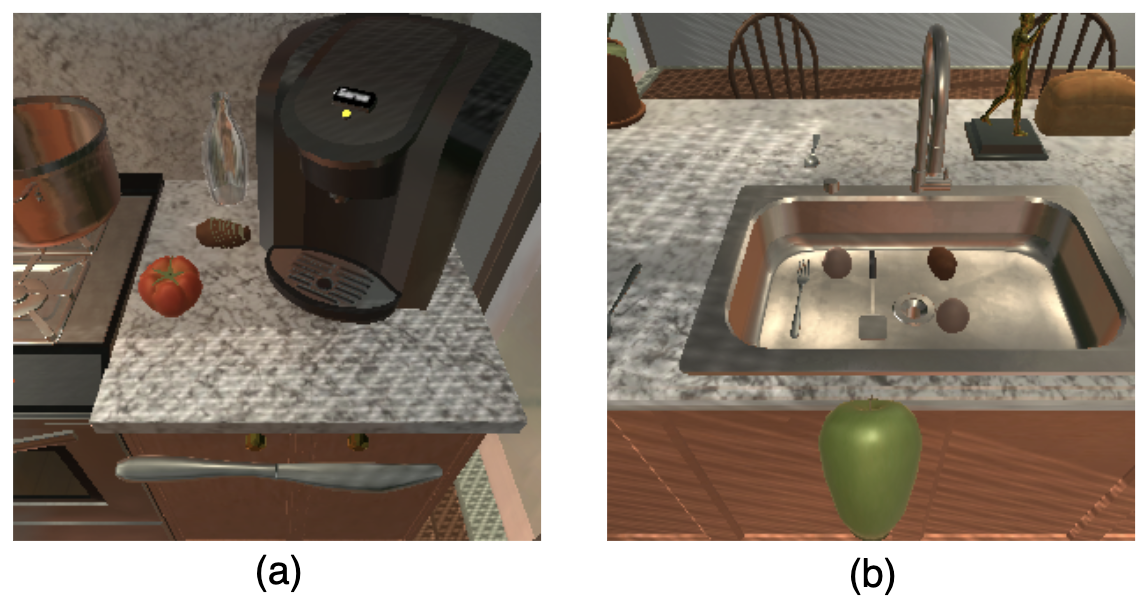}
    \vspace{-7mm}
    \caption{\small Successful subgoal predictions. (a) ``Place a cooked potato slice in the fridge.'' (b) ``Get an apple from the sink and heat it up in the microwave.''}
    \label{fig:subgoals}
\end{figure}

\subsection{Ablation Studies}
We set the following three ablation conditions:
\begin{enumerate}
    \renewcommand{\labelenumi}{(\roman{enumi})}
    \item W/o MAT: We removed MAT to investigate the effect of the proposed moment update of the perturbations. We used the procedure proposed in VILLA\cite{gan2020large} instead.
    \item Selective adversarial training: We did not add perturbations to all the input: instruction, subgoals, and state representations. We selected two of them to apply adversarial training to clarify which inputs were effective when we added perturbations to them.
    \item More/fewer adversarial training steps: We increased or decreased the adversarial training steps to investigate the effect of the number of perturbation updates. We set the adversarial training step to 3, 5, 9, or 11.
\end{enumerate}

Table~\ref{tab:quantitative_results} shows the quantitative results of the ablation study. On the test set, the Unseen SR was 20.99\% without MAT. Additionally, the Unseen SR was 21.39\%, 21.52\%, and 20.27\% without perturbations added to the instruction, subgoals, and state representations, respectively. This indicates that MAT w.r.t. state representations has the biggest influence on performance.

Fig.~\ref{fig:graph} shows the Unseen SRs of our method for different adversarial training steps. It shows that our method achieved the highest score for an adversarial training step of 7 on both the validation and test sets.

\subsection{Qualitative Results}
Figs.~\ref{fig:success_01} and \ref{fig:success_02} show success samples. Fig.~\ref{fig:success_01} illustrates that the robot correctly took a cup to a lamp and turned it on. Similarly, Fig.~\ref{fig:success_02} shows that the robot correctly carried the pillow to the sofa twice.

Subgoal-level results are shown below.
Fig.~\ref{fig:subgoals} (a) shows a correct prediction. The instruction was ``Place a cooked potato slice in the fridge.'' At this timestep, the robot was required to put the butter knife on the counter because it had already been used to slice the potato and the robot had to carry the sliced potato to the fridge. The baseline method output the action: ``Open,'' whereas our method correctly output the required action. This indicates that our method was able to predict the unspecified and necessary action: ``release the knife.''
Fig.~\ref{fig:subgoals} (b) shows another correct prediction. The instruction was ``Get an apple from the sink and heat it up in the microwave.'' Fig.~\ref{fig:subgoals} (b) shows the scenario after the apple was heated in the microwave, and the remaining task was to put it in the sink. The baseline method output the action: ``Open,'' whereas our method correctly output the required action. As in the above sample, this indicates that our method correctly estimated the unspecified subgoal: ``place the apple.''

\subsection{Error Analysis and Discussion}
Fig.~\ref{fig:failure_01} shows a failure sample. The instruction was ``Put a cooled potato inside the microwave.'' The robot correctly picked up a potato and put it in the fridge initially. Then, the robot mistakenly picked up another potato while it should have taken the cooled one.

Table~\ref{tab:f1_list} presents the number of ground truth samples and F1-scores corresponding to each subgoal type.
Table~\ref{tab:f1_list} shows that the F1-score of ``Slice'' was particularly low. This is because the number of ``Slice'' subgoals in the training set was fewer than others.

\begin{table}[b]
    \small
    \vspace{-5mm}
    \caption{GT samples and F1-scores of subgoal types}
    \vspace{-2mm}
    \label{tab:f1_list}
    \centering
    \begin{tabular}{l c c}
        \hline
        Subgoal type & \#GT samples & F1 \\
        \hline
        PickUp & 1348 & 0.963 \\
        Put & 1202 & 0.963 \\
        Open & 831 & 0.905 \\
        Close & 846 & 0.998 \\
        ToggleOn & 361 & 0.981 \\
        ToggleOff & 221 & 0.998 \\
        Slice & 110 & 0.890 \\
        \hline
    \end{tabular}
\end{table}

\section{Conclusion}
\label{sec:conclusion}

In this paper, we focused on a multimodal language comprehension model that translates from language to sequences of actions.
We would like to emphasize the following contributions of this study:
\begin{itemize}
    \item We introduced adversarial perturbations to the embedding spaces of subgoals and state representations.
    \item We proposed MAT, which is a new algorithm that uses two types of moments for perturbation updates.
    \item Our method outperformed the baseline method for all the metrics on the ALFRED benchmark\cite{shridhar2020alfred}.
\end{itemize}


\section*{ACKNOWLEDGEMENT}
This work was partially supported by JSPS KAKENHI Grant Number 20H04269, JST Moonshot, and NEDO.

\bibliographystyle{IEEEtran}
\bibliography{reference}

\end{document}